\documentclass[runningheads]{llncs}

\usepackage[T1]{fontenc}
\usepackage{graphicx}
\usepackage{booktabs}
\usepackage{amsmath}
\usepackage{amssymb}
\usepackage{multirow}
\usepackage{xcolor}
\usepackage{hyperref}
\usepackage{cleveref}
\usepackage[normalem]{ulem}
\usepackage{tikz}
\usetikzlibrary{positioning,shapes,arrows.meta,fit,backgrounds}

\begin{document}

\title{Knowledge-Enriched Structured EHR Features for 30-Day Hospital Readmission Prediction on MIMIC-IV}
\titlerunning{Knowledge-Enriched Structured EHR for Readmission Prediction}

\author{Mohamad Najafi\inst{1} \and Hongyun Fu\inst{1} \and Mathias Brochhausen\inst{2} \and Jian Wu\inst{1}\thanks{Corresponding author.} \and Yaohang Li\inst{1}\thanks{Corresponding author.}}
\authorrunning{M. Najafi et al.}
\institute{Old Dominion University, Virginia, USA%
\and University of Arkansas for Medical Sciences, Arkansas, USA%
}

\maketitle

\begin{abstract}
Recent approaches to 30-day hospital readmission prediction rely on pre-trained language models applied to discharge summaries. Although these methods achieve strong performance, they depend on the availability of clinical notes, incur substantial computational costs, and yield representations that lack interpretability.
We propose a knowledge-enriched feature representation that augments structured Electronic Health Record (EHR) data with four medical knowledge sources: disease ontology mapping, procedure classification, drug ingredient vocabulary, and organ system laboratory aggregation, without using clinical notes.
Each feature dimension corresponds to a named clinical concept, yielding a sparse and interpretable patient representation.
The approach is evaluated with six classifiers on a MIMIC-IV v2.2 cohort.
Under 20-fold cross-validation, the best configuration achieves an AUROC of $0.743$. This performance is comparable to that of previously reported methods on this dataset---including both those using only structured data and those incorporating clinical notes---while requiring considerably less computational cost.
Interpretability analysis shows that demographics, organ system labs, drug ingredient features, and first-level ontology disease categories drive prediction, while deeper hierarchy levels contribute negligibly.
These findings indicate that knowledge-enriched structured features offer a competitive and efficient alternative to embeddings from clinical notes for 30-day readmission prediction.

\end{abstract}


\section{Introduction}\label{sec:introduction}

Unplanned 30-day hospital readmissions remain a persistent quality and cost challenge in healthcare systems.
In the United States, the Hospital Readmissions Reduction Program (HRRP) imposes financial penalties on hospitals with excess readmission rates \cite{muchiri2022analysis}, and readmission rates serve as a widely adopted indicator of care quality across countries \cite{van2010derivation}.
Accurate prediction of readmission risk at the point of discharge enables targeted interventions, such as transitional care programs and structured outpatient follow-up \cite{balasubramanian2025outpatient}, which have proven effective at reducing avoidable readmissions \cite{pattar2025electronic}.

Recently, several methods have incorporated unstructured clinical text to improve readmission prediction.
Almeida et al.\ \cite{almeida2025prediction} achieved an area under the receiver operating characteristic curve (AUROC $\in [0,1]$) of $0.727$ on MIMIC-IV \cite{johnson2023mimic} by encoding discharge summaries with a pre-trained language model and combining the resulting embeddings with structured features in a graph neural network.
Other text-based and multimodal approaches, including MM-STGNN \cite{tang2023predicting} and a ClinicalT5-based model \cite{pandey2025predicting}, report comparable AUROC values on restricted MIMIC-IV subsets.

Structured Electronic Health Record (EHR) data, by contrast, are available for all billed admissions and, because they are already recorded as discrete coded fields (diagnoses, procedures, laboratory results, and medications), require no natural language processing to encode \cite{rajkomar2018scalable}.
In this work, we focus on diagnosis codes, procedure codes, laboratory results, and medication records, which cover the principal clinical drivers of readmission risk (comorbidity, interventions, physiological status, and treatment) and are populated for essentially all billed admissions in MIMIC-IV. Other structured signals (e.g., vital signs, microbiology) are sparser or less standardized and are left to future work.
However, raw structured features encode clinical events as flat categorical variables without capturing the semantic relationships between them \cite{choi2017gram}.
For instance, ICD-10 E11.9 (type~2 diabetes without complications) and E11.65 (type~2 diabetes with hyperglycemia) share no explicit similarity unless an external ontology links them through a common disease ancestor; Makohon et al.\ \cite{makohon2025enhancing} showed that mapping ICD codes to SNOMED CT concepts via ontological definitions establishes such clinically meaningful links.
Medical ontologies, including the Mondo Disease Ontology (MONDO) \cite{vasilevsky2022mondo}, the Clinical Classifications Software (CCS) \cite{salsabili2020evaluation}, and the RxNorm drug terminology \cite{nelson2011normalized}, organize clinical concepts into hierarchical structures that encode precisely these relationships.

This paper investigates whether augmenting structured EHR features with knowledge derived from four medical sources produces a readmission prediction model that is competitive with note-based approaches.
The contributions are: (1)~a knowledge-enriched feature representation integrating four medical knowledge sources with structured EHR data, producing a sparse patient representation in which each dimension corresponds to a named clinical concept; (2)~a controlled three-way comparison of ontology-enriched, transformer-encoded, and combined features on a MIMIC-IV v2.2 cohort with six classifiers; (3)~an interpretability analysis quantifying the contribution of each knowledge source; and (4)~evidence that knowledge-enriched structured features achieve competitive 30-day readmission prediction without clinical notes.


\section{Related Work}\label{sec:related_work}

\subsection{Readmission Prediction from Structured EHR}\label{sec:rw_structured}

Early readmission risk models relied on manually curated scoring systems.
The LACE index \cite{van2010derivation} combines length of stay, acuity, comorbidity, and ED visits into an additive score (AUROC ${\sim}$0.68), and the HOSPITAL score \cite{donze2016international} uses seven structured variables with similar performance.
Machine learning approaches extended these baselines: Rajkomar et al.\ \cite{rajkomar2018scalable} applied deep learning to EHR from 216{,}221 patients, achieving AUROC 0.75--0.76 for 30-day readmission; Ashfaq et al.\ \cite{ashfaq2019readmission} achieved AUROC 0.77 with a cost-sensitive LSTM for congestive heart failure (CHF) readmission.

\subsection{Note-Based and Multimodal Readmission Prediction}\label{sec:rw_notes}

Beyond structured features, pre-trained language models applied to clinical text have improved readmission prediction.
Huang et al.\ \cite{huang2019clinicalbert} fine-tuned ClinicalBERT for discharge-based readmission prediction and demonstrated that clinical notes carry complementary signal beyond structured codes.
Almeida et al.\ \cite{almeida2025prediction} combined BioClinicalBERT \cite{alsentzer2019publicly} embeddings of diagnosis and procedure titles and discharge summaries with a Facebook AI Similarity Search (FAISS) \cite{johnson2019billion} patient graph and a GraphSAGE \cite{hamilton2017inductive} classifier, achieving AUROC 0.727 on MIMIC-IV \cite{johnson2023mimic} (structured-only baseline 0.704).
Tang et al.\ \cite{tang2023predicting} proposed MM-STGNN, a spatiotemporal graph neural network fusing longitudinal chest radiographs and structured EHR data, achieving AUROC 0.79 on a chest-radiograph subset of MIMIC-IV.
Pandey et al.\ \cite{pandey2025predicting} applied a fine-tuned ClinicalT5 model to clinical notes combined with structured features, reporting AUROC 0.68 on a discharge-note subset of MIMIC-IV.
He et al.\ \cite{he2025comparative} compared structured features against discharge narratives on MIMIC-IV (AUROC 0.65--0.67 for classical ML vs.\ 0.72 for ClinicalLongformer), and Shakya et al.\ \cite{shakya2025predicting} found word2vec EHR embeddings outperformed one-hot and pre-trained BERT for heart-failure readmission.
\subsection{Knowledge-Enriched Clinical Prediction}\label{sec:rw_knowledge}

An alternative to note-based encoding is integrating external medical knowledge directly into structured representations, avoiding the dependency on discharge summary availability.
Choi et al.\ \cite{choi2017gram} proposed GRAM, which learns medical concept representations by attending over ancestors in a medical ontology, and applied it to diagnosis prediction on MIMIC-III \cite{johnson2016mimic}.
Rasmy et al.\ \cite{rasmy2021med} pretrained Med-BERT on 28.5 million patient records, improving AUROC by 1.21--6.14\% over base RNN models (GRU, Bi-GRU, RETAIN) on two disease-prediction tasks.
Xu et al.\ \cite{xu2024ram} developed RAM-EHR, achieving a 3.4\% average AUROC gain over prior knowledge-enhanced baselines on phenotype and cardiovascular-outcome prediction.
Jiang et al.\ \cite{jiang2024reasoning} proposed KARE, a KG community retrieval framework that improves MIMIC-IV readmission prediction by up to 12.7\% in macro-F1 over the best baseline.
Carvalho et al.\ \cite{carvalho2023knowledge} enriched MIMIC-III ICU stays with annotations from seven biomedical ontologies and learned dense KG embeddings (AUROC 0.827 for ICU readmission); their approach produces opaque learned vectors rather than explicit interpretable features and targets ICU rather than all-cause hospital readmission.
These works establish the value of ontological structure for clinical modeling but have not applied multi-source knowledge enrichment as explicit sparse features to structured-only hospital readmission prediction, nor compared it against transformer-based feature extraction on an identical cohort.


\section{Problem Formulation and Study Motivation}\label{sec:problem}

Note-based models can improve discrimination but depend on discharge summaries, which are not recorded for every admission and are costly to process, whereas structured-only models avoid this dependency at a possible accuracy cost. This trade-off motivates the structured-only setting we study next.

We address the standard 30-day readmission prediction task under a structured-only input setting; the task is unchanged from prior work, and only the input representation varies.
Formally, it is a binary classification problem: given an admission $a_i$ with structured feature vector $\mathbf{x}_i$, the goal is to predict $y_i \in \{0, 1\}$, where $y_i = 1$ if and only if the same patient has a subsequent admission within 30 calendar days of discharge from $a_i$.
Because the instances (admissions) and target are identical to prior work, AUROC values remain directly comparable across feature configurations.
Each admission is treated independently; we do not model temporal sequences across admissions.

To investigate the role of feature representation, we define three configurations evaluated under identical conditions, with each component detailed in Section~\ref{sec:framework}:

\begin{itemize}
\item \textbf{B1 (BERT-based)}: BioClinicalBERT \cite{alsentzer2019publicly} embeddings of ICD diagnosis and procedure \emph{titles}, concatenated with raw demographic features and per-item lab abnormality rates.
\item \textbf{B2 (Ontology-enriched)}: demographic features concatenated with multi-level MONDO disease-hierarchy features (Section~\ref{sec:mondo}), CCS procedure categories, organ-system laboratory aggregates, RxNorm drug-ingredient indicators, and unmapped-ICD features.
\item \textbf{B3 (Combined)}: B1 reduced to 256 dimensions by truncated SVD (Section~\ref{sec:integration}) concatenated with the full B2 matrix.
\end{itemize}

B1 is a transformer-encoded representation of structured features; B2 replaces these learned embeddings with explicit knowledge-enriched features; B3 tests complementarity.
B2 is sparse, requiring approximately 24 times less storage than B1 (176\,MB versus 4.2\,GB for the full cohort), and each dimension maps to a named clinical concept.


\section{Dataset and Cohort Construction}\label{sec:dataset}

We use MIMIC-IV v2.2 \cite{johnson2023mimic}, a de-identified single-center EHR dataset (2008--2019), extracting structured data from the \texttt{admissions}, \texttt{patients}, \texttt{diagnoses\_icd}, \texttt{procedures\_icd}, \texttt{prescriptions}, and \texttt{labevents} tables; diagnosis codes span ICD-9-CM and ICD-10-CM.

\paragraph{Cohort construction.}
Starting from all admissions, we exclude patients under 18 and in-hospital deaths, yielding 422{,}622 admissions across 176{,}901 patients.
We impose no note-related inclusion criteria, so the cohort is not restricted to admissions with available discharge summaries.

\paragraph{Label definition.}
Following the definition in Section~\ref{sec:problem}, the chronologically last admission of each patient has no subsequent admission on record and therefore cannot be a 30-day readmission, so it receives $y_i = 0$.
The 30-day readmission rate is 20.1\%.

\paragraph{Data split.}
We partition the data into training (60\%), validation (20\%), and test (20\%) sets at the \emph{patient level}: all admissions of a given patient are assigned to a single split, so no patient appears in more than one split and no information leaks between training and test.
The split is stratified by each patient's maximum label, preserving the readmission ratio across splits.
The cohort has mean age $56.6 \pm 19.0$ years and is 52.3\% female; it is split into 253{,}254 training, 84{,}328 validation, and 85{,}040 test admissions, and contains 25{,}809 unique ICD diagnosis codes, 12{,}575 procedure codes, and 9{,}610 prescription drug names.


\section{Knowledge-Enriched Readmission Prediction Framework}\label{sec:framework}

\begin{figure}[t]
\centering
\resizebox{0.86\textwidth}{!}{%
\begin{tikzpicture}[
  >=Stealth,
  every node/.style={font=\small},
  ehrbox/.style={draw=blue!50, fill=blue!10, rounded corners=5pt,
                 align=center, text width=2.6cm, minimum height=4.2cm},
  integbox/.style={draw=green!50!black, fill=green!10, rounded corners=5pt,
                   align=center, text width=2.8cm, minimum height=4.2cm},
  knowbox/.style={draw=orange!60, fill=yellow!20, rounded corners=4pt,
                  align=center, text width=1.75cm, minimum height=1.2cm,
                  font=\footnotesize},
  matbox/.style={draw=orange!70, fill=orange!15, rounded corners=5pt,
                 align=center, text width=2.4cm, minimum height=3.2cm},
  treebox/.style={draw=blue!60, fill=blue!15, rounded corners=5pt,
                  align=center, text width=2.5cm, minimum height=1.6cm},
  svdbox/.style={draw=violet!60, fill=violet!10, rounded corners=5pt,
                 align=center, text width=2.5cm, minimum height=1.2cm},
  faisbox/.style={draw=red!50, fill=pink!20, rounded corners=5pt,
                  align=center, text width=2.5cm, minimum height=1.2cm},
  neubox/.style={draw=green!60!black, fill=green!10, rounded corners=5pt,
                 align=center, text width=2.5cm, minimum height=1.6cm},
  resbox/.style={draw=green!60!black, fill=green!25, rounded corners=5pt,
                 align=center, text width=2.5cm, minimum height=3.2cm},
  lbl/.style={font=\scriptsize\itshape, text=black},
]

\node[ehrbox] (ehr) at (0,0) {
  \textbf{Structured EHR}\\[4pt]
  Raw MIMIC-IV\\Evidence Data\\[4pt]
  Admissions\\Diagnoses (ICD)\\Procedures\\Lab events\\Prescriptions
};

\node[integbox] (integ) at (4.0,0) {
  \textbf{Knowledge}\\
  \textbf{Integration}\\[4pt]
  Ontology-based\\mapping and\\enrichment
};

\node[knowbox] (mondo) at (3.0,-3.6) {MONDO\\Disease\\Ontology};
\node[knowbox, right=0.3cm of mondo] (ahrq)   {AHRQ\\CCS};
\node[knowbox, left=0.3cm of mondo] (organ)  {Organ-sys.\\Lab Map};
\node[knowbox, right=0.3cm of ahrq] (rxnorm) {RxNorm};

\node[font=\footnotesize\bfseries\itshape, text=orange!80!black]
  at (4.0,-4.65) {4 Knowledge Sources};

\node[matbox] (b2) at (8.0,0) {
  \textbf{Feature}\\
  \textbf{Matrix B2}\\[4pt]
  23{,}430 dims\\99.78\% sparse\\
};

\node[treebox] (tree) at (11.8, 1.9) {
  \textbf{Tree / Linear}\\
  \textbf{Models}\\(full sparse)
};
\node[lbl, above=1.5mm of tree] {LightGBM, XGBoost, LR, RF};

\node[svdbox]  (svd)   at (11.8, 0.1)  {Truncated SVD\\$k$=2{,}048 (95.4\% var)};
\node[faisbox] (faiss) at (11.8,-1.4)  {FAISS Graph\\cosine\,$>$\,0.9,\,${\sim}$17M edges};
\node[neubox]  (neural) at (11.8,-3.1) {
  \textbf{Neural Models}
};
\node[lbl, below=1.5mm of neural] {MLP, GraphSAGE};

\node[resbox] (result) at (15.4,-0.6) {
  \textbf{Best AUROC}\\
  \textbf{0.7412}\\[4pt]
  B2 LightGBM\\(structured only)
};

\draw[->]              (ehr)   -- (integ);
\draw[->]              (integ) -- (b2);

\draw[->, orange!70] (mondo.north)  to[out=90,in=-90] (integ.south);
\draw[->, orange!70] (ahrq.north)   to[out=90,in=270] (integ.south);
\draw[->, orange!70] (organ.north)  to[out=90,in=260] (integ.south);
\draw[->, orange!70] (rxnorm.north)  to[out=90,  in=280] (integ.south);

\draw[->, blue!60]
  (b2.east) to[out=30,in=180]
  node[lbl, above=1pt] {direct} (tree.west);

\draw[->, violet!60]
  (b2.east) to[out=-15,in=180]
  node[lbl, below=1pt] {project} (svd.west);

\draw[->, violet!60]  (svd)   -- (faiss);
\draw[->, green!60!black] (faiss) -- (neural);

\draw[->, blue!60]
  (tree.east) to[out=0,in=90] (result.north);

\draw[->, green!60!black]
  (neural.east) to[out=0,in=270] (result.south);

\end{tikzpicture}%
}
\caption{Knowledge-enriched feature construction pipeline. Raw MIMIC-IV structured data is enriched through four medical knowledge sources via ontology-based mapping into the sparse B2 feature matrix. Tree and linear models operate directly on the full sparse matrix; neural models use SVD-2048 projections.}
\label{fig:pipeline}
\end{figure}
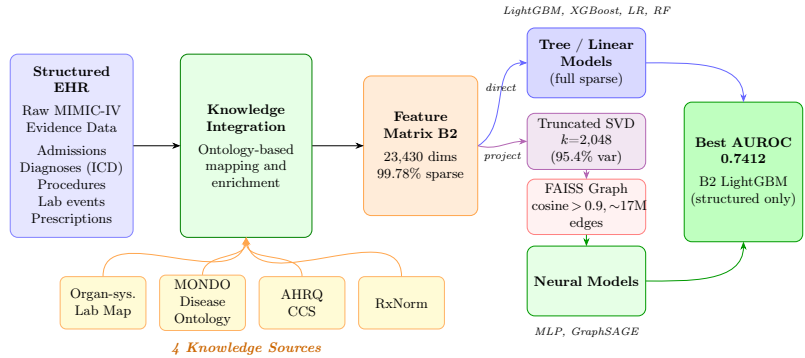

The B2 ontology-enriched features are built from the same MIMIC-IV tables, mapping raw structured data through four knowledge sources (MONDO, CCS, RxNorm, and organ-system laboratory aggregation) into six feature blocks (demographics, MONDO disease hierarchy, CCS procedures, organ-system labs, RxNorm drugs, and unmapped ICD codes), concatenated into a single sparse matrix; the ablation (Section~\ref{sec:ablation}) reports the MONDO block at its three ancestor levels (L1--L3), giving eight rows.
Figure~\ref{fig:pipeline} provides an overview.

\subsection{Demographics}\label{sec:demographics}

The demographic and administrative block encodes administrative fields (admission type, admission location, discharge location, insurance type) and demographic fields (race/ethnicity, marital status) as one-hot vectors, plus age at admission and length of stay as min-max scaled continuous values, yielding 93 features shared across B1, B2, and B3.

\subsection{MONDO Disease Ontology Mapping}\label{sec:mondo}

The Mondo Disease Ontology (MONDO) \cite{vasilevsky2022mondo} provides a unified disease classification integrating OMIM \cite{amberger2015omim}, Orphanet \cite{rath2012representation}, and other sources through the Open Biological and Biomedical Ontologies (OBO) Foundry framework \cite{jackson2021obo}.
Following the preprocessing of Almeida et al.\ \cite{almeida2025prediction}, we retain the first 10 diagnosis codes per admission (ordered by sequence number), which keeps the structured diagnosis inputs consistent with the B1 comparator; we then map each ICD-9 or ICD-10 code to MONDO concepts via cross-reference (\texttt{xref}) links, attempting code truncation at three levels when an exact match fails.
Here a MONDO \emph{concept} denotes a disease entity (a node in the ontology) to which a diagnosis code is mapped.

\begin{figure}[h]
\centering
\resizebox{0.92\textwidth}{!}{%
\begin{tikzpicture}[
  >=Stealth,
  every node/.style={font=\tiny},
  l3box/.style={draw=green!60!black, fill=green!45, rounded corners=3pt,
                align=center, text width=2.0cm, minimum height=1.0cm},
  l2box/.style={draw=green!50!black, fill=green!20, rounded corners=3pt,
                align=center, text width=2.0cm, minimum height=1.0cm},
  l1box/.style={draw=orange!70, fill=orange!20, rounded corners=3pt,
                align=center, text width=2.0cm, minimum height=1.0cm},
  icdbox/.style={draw=blue!60, fill=blue!15, rounded corners=3pt,
                 align=center, text width=2.1cm, minimum height=0.9cm},
  arr/.style={->, thick, gray!60},
  legsq/.style={draw, minimum size=0.28cm, inner sep=0pt},
]

\node[l3box] (l3) {
  cardiovascular\\disease\\(MONDO:0004995)\\[1pt]\textit{\tiny L3}
};

\node[l2box, right=0.55cm of l3] (l2) {
  heart\\disease\\(MONDO:0005267)\\[1pt]\textit{\tiny L2}
};

\node[l1box, right=0.55cm of l2] (l1) {
  congestive\\heart failure\\(MONDO:0005252)\\[1pt]\textit{\tiny L1}
};

\node[icdbox, above right=0.35cm and 0.55cm of l1] (icd10) {
  ICD-10 I50.9\\{\tiny(heart failure)}
};

\node[icdbox, below right=0.35cm and 0.55cm of l1] (icd9) {
  ICD-9 428.0\\{\tiny(CHF unspec.)}
};

\draw[arr] (l3) -- (l2);
\draw[arr] (l2) -- (l1);
\draw[arr] (l1.east) to[out=30, in=180]  (icd10.west);
\draw[arr] (l1.east) to[out=-30,in=180]  (icd9.west);

\node[legsq, fill=green!45, draw=green!60!black] (lsq3) at (-1,-2) {};
\node[font=\tiny, right=1.2mm of lsq3, anchor=west] {L3: great-grandparent};

\node[legsq, fill=green!20, draw=green!50!black, right=3.4cm of lsq3] (lsq2) {};
\node[font=\tiny, right=1.2mm of lsq2, anchor=west] {L2: grandparent};

\node[legsq, fill=orange!20, draw=orange!70, right=2.8cm of lsq2] (lsq1) {};
\node[font=\tiny, right=1.2mm of lsq1, anchor=west] {L1: direct parent};

\node[legsq, fill=blue!15, draw=blue!60, right=2.6cm of lsq1] (lsqicd) {};
\node[font=\tiny, right=1.2mm of lsqicd, anchor=west] {Raw ICD-9/ICD-10};

\end{tikzpicture}%
}
\caption{MONDO disease ontology mapping (cardiovascular example). ICD codes are linked to three ancestor levels (L1--L3); a binary feature is generated for each mapped concept.}\label{fig:mondo}
\end{figure}
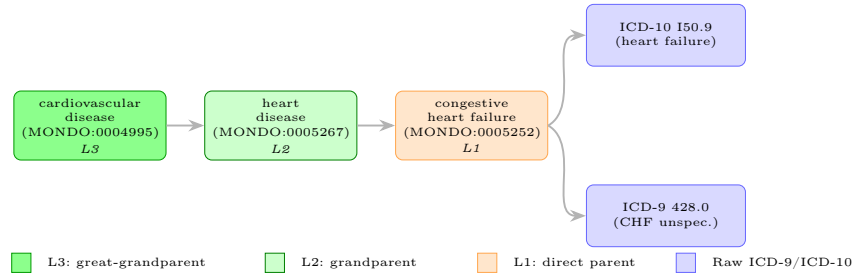

For each mapped concept, three ancestor levels are extracted (L1: direct parent, L2: grandparent, L3: great-grandparent), as illustrated in Figure~\ref{fig:mondo}.
This yields 5{,}413 L1, 892 L2, and 252 L3 binary features (6{,}557 mapped features in total).
ICD codes with no cross-reference to any MONDO concept, typically administrative, symptom, or factor codes (e.g., ICD-10 Z-chapter ``factors influencing health status'' codes), are termed \emph{unmapped} and retained as raw one-hot features in a separate block of 14{,}905 dimensions.
Diagnosis-derived features thus total 21{,}462 dimensions ($6{,}557 + 14{,}905$); with the remaining blocks the full B2 matrix spans 23{,}430 (Section~\ref{sec:integration}).

\subsection{CCS Procedure Classification}\label{sec:ccs}

Raw ICD procedure codes are high-cardinality and sparse. The AHRQ Clinical Classifications Software (CCS) \cite{salsabili2020evaluation} groups them into mutually exclusive, clinically coherent categories, reducing dimensionality while preserving procedural meaning; in our cohort it maps all 12{,}575 unique procedure codes into 527 single-level categories.
As in Almeida et al.\ \cite{almeida2025prediction}, we map the first five procedures per admission (by sequence number) to these categories as a multi-hot vector of 527 binary features; admissions with fewer than five yield a sparser vector with no padding.

\subsection{Organ system Laboratory Aggregation}\label{sec:labs}

Laboratory results from \texttt{labevents} are grouped into eight categories by keyword matching against MIMIC-IV's \texttt{d\_labitems} table \cite{johnson2023mimic}: cardiac, hepatic, renal, hematologic, metabolic, inflammatory, coagulation, and a residual category for unmatched tests.
For example, serum troponin-I results are assigned to the cardiac system and creatinine results to the renal system.
For each system, three features are computed (total count, abnormal count, abnormal percentage), yielding $8 \times 3 = 24$ continuous features.
A \emph{system} here is an organ or physiological grouping of tests, distinct from the CCS procedure categories; an admission with no results for a system has those three features set to zero (no imputation).

\subsection{RxNorm Drug Ingredient Mapping}\label{sec:drugs}

Prescription free-text drug names are normalized and matched against the RxNorm CUI vocabulary \cite{nelson2011normalized}.
Each matched term is collapsed to its ingredient-level RxCUI by following \texttt{has\_ingredient}, \texttt{tradename\_of}, and \texttt{consists\_of} relations in the RxNorm relationship file.
For example, ``metformin 500\,mg'' normalizes and collapses to RxCUI~41493 (metformin, ingredient).
Not every drug matches: 72.1\% of prescription records (41.1\% of unique strings) map to an ingredient RxCUI, and unmatched drugs are dropped.
Each unique ingredient CUI present in an admission becomes a binary feature, yielding 1{,}324 dimensions.

\subsection{Feature Integration}\label{sec:integration}

The six blocks are concatenated via sparse horizontal stacking into a CSR matrix of 23{,}430 dimensions.
For neural models (MLP, GraphSAGE), the sparse matrix is projected to a dense representation via truncated singular value decomposition (SVD) \cite{halko2011finding} with $k=2{,}048$ components (95.4\% variance retained).
Tree and linear models operate directly on the full sparse matrix.

B3 projects B1 (2{,}481 dimensions) to 256 via truncated SVD, then concatenates with full B2, yielding 23{,}686 dimensions for tree/linear models; neural models use SVD-2048 of B3 (95.2\% variance retained).
Preliminary experiments showed that SVD-256 retains only 71\% of variance whereas SVD-2048 captures over 95\%, yielding approximately two AUROC points improvement for neural models.
Neural message-passing models can exploit relational structure between clinically similar patients, which a per-admission feature vector does not represent. We therefore construct a patient graph from the dense embeddings: indexing them with FAISS \cite{johnson2019billion}, we retrieve neighbors by exact inner-product (cosine) search and link patients with cosine similarity $> 0.9$ by an edge (17{,}401{,}420 edges over 422{,}622 patients). The resulting graph is passed to GraphSAGE for message passing.


\section{Experimental Setup}\label{sec:experiments}

\paragraph{Models.}
We evaluate six classifiers: LightGBM \cite{ke2017lightgbm} (31 leaves, learning rate 0.05, 300 trees), XGBoost \cite{chen2016xgboost} (depth 6, learning rate 0.1, 300 trees), L2-regularized Logistic Regression (LR, $C=1$), Random Forest \cite{breiman2001random} (RF, 200 trees, depth 20, min.\ leaf 5), a Multilayer Perceptron (MLP, hidden $512$--$256$--$128$, dropout 0.3), and GraphSAGE \cite{hamilton2017inductive} (two SAGE layers, hidden 64, neighbors $[10,10]$). The two neural models use Adam (MLP learning rate $5 \times 10^{-4}$, GraphSAGE $10^{-5}$), batch 1024, and up to 150 epochs with early stopping (patience 10).

Each model is evaluated on all three feature sets (B1, B2, B3): tree and linear models (LightGBM, XGBoost, LR, RF) consume the full sparse matrix, the neural models (MLP, GraphSAGE) its SVD-2048 projection (Table~\ref{tab:features}, Section~\ref{sec:integration}).

\paragraph{Evaluation metrics.}
The primary metric is AUROC.
The secondary metric is Balanced Accuracy (BAcc), both reported for the held-out test set across all models.
AUPRC is additionally reported for the best-performing model (LightGBM on B2) to complement AUROC under class imbalance.
The label is imbalanced (20.1\% positive); all models address this via class weighting (balanced class weights for LR and RF, \texttt{scale\_pos\_weight} for the boosted trees, and a positive-weighted loss for the neural models).
The classification threshold is selected on the validation set by maximizing balanced accuracy.
Although the cohort is large, we report 20-fold cross-validation for the best model (LightGBM on B2) to quantify estimate variance and match the protocol of prior work \cite{almeida2025prediction}; its mean and 95\% CI are reported in Table~\ref{tab:results}.

\paragraph{Feature configurations.}
Table~\ref{tab:features} summarizes the three feature sets; all share the same cohort, label definition, and patient-level 60/20/20 split.

\begin{table}[t]
\caption{Feature set configurations. B1 follows the pipeline of Almeida et al.\ \cite{almeida2025prediction}; B2 is the proposed knowledge-enriched representation; B3 combines both. Neural models use SVD-2048 projections for B2 and B3.}\label{tab:features}
\centering
\begin{tabular}{llrr}
\toprule
\textbf{Set} & \textbf{Components} & \textbf{Dims} & \textbf{Sparsity} \\
\midrule
B1 & Demo + BERT diag/proc + labs & 2{,}481 & Dense \\
B2 & Demo + MONDO + CCS + labs + drugs + ICD & 23{,}430 & 99.78\% \\
B3 & B1(SVD-256) + B2 & 23{,}686 & Mixed \\
\bottomrule
\end{tabular}
\end{table}

\paragraph{Hardware.}
Experiments were conducted on a workstation with an Intel Core Ultra~9 285 (24 cores), 64\,GB RAM, and an NVIDIA RTX~4000 SFF Ada (20\,GB).


\section{Results}\label{sec:results}

Table~\ref{tab:results} reports AUROC and Balanced Accuracy for all six classifiers across the three feature configurations on the held-out test set.

\begin{table}[t]
\caption{Held-out fixed-split test AUROC and Balanced Accuracy (BAcc) for six models across three feature configurations. \textbf{Bold}: best feature set per model. \underline{Underline}: best model per feature set. Under 20-fold cross-validation at the patient level, the best configuration (LightGBM, B2) achieves AUROC $0.743 \pm 0.007$ (BAcc 0.676). For reference, on a single split Almeida et al.\ \cite{almeida2025prediction} report AUROC 0.704 (structured-only) and 0.727 (with notes), with BAcc 0.649 and 0.667 respectively, on a cohort conditioned on discharge-summary availability, with different inclusion criteria.}\label{tab:results}
\centering
\begin{tabular}{l cc cc cc}
\toprule
 & \multicolumn{2}{c}{\textbf{B1 (BERT)}} & \multicolumn{2}{c}{\textbf{B2 (Ontology)}} & \multicolumn{2}{c}{\textbf{B3 (Combined)}} \\
\cmidrule(lr){2-3} \cmidrule(lr){4-5} \cmidrule(lr){6-7}
\textbf{Model} & AUROC & BAcc & AUROC & BAcc & AUROC & BAcc \\
\midrule
LightGBM  & \underline{0.7359} & \underline{0.6726} & \underline{\textbf{0.7412}} & \underline{0.6739} & \underline{0.7396} & \underline{0.6769} \\
XGBoost   & 0.7322 & 0.6674 & \textbf{0.7374} & 0.6714 & 0.7359 & 0.6732 \\
MLP       & 0.7323 & 0.6701 & 0.7317 & 0.6685 & \textbf{0.7325} & 0.6693 \\
GraphSAGE & \textbf{0.7271} & 0.6679 & 0.7239 & 0.6626 & 0.7260 & 0.6627 \\
LR        & \textbf{0.7164} & 0.6602 & 0.7049 & 0.6538 & 0.7100 & 0.6564 \\
RF        & 0.7007 & 0.6450 & 0.6966 & 0.6421 & \textbf{0.7109} & 0.6552 \\
\bottomrule
\end{tabular}
\end{table}

\paragraph{Feature representation and model comparison.}
B2 (ontology-enriched) achieves the highest AUROC for gradient-boosted tree models: LightGBM reaches 0.7412 and XGBoost 0.7374, both significantly outperforming their B1 counterparts (bootstrap, 1{,}000 iterations; LightGBM: mean $\Delta = +0.028$, 95\% CI [+0.026, +0.031], $p < 0.001$; LR: mean $\Delta = +0.010$, 95\% CI [+0.006, +0.014], $p < 0.001$). We report the bootstrap test for LightGBM (best tree model) and LR (linear baseline) as family representatives; the other models follow the same B2-versus-B1 direction (Table~\ref{tab:results}).
Dense-input models show the opposite pattern: GraphSAGE (0.7271 on B1 vs.\ 0.7239 on B2) and LR (0.7164 vs.\ 0.7049) prefer B1.
B3 does not consistently outperform either single-source configuration.
LightGBM on B2 is the best result across all 18 configurations (AUROC 0.7412, AUPRC 0.441).

\paragraph{Cross-study context.}
The best fixed-split result (LightGBM B2, AUROC 0.7412), and the 20-fold-CV value of $0.743 \pm 0.007$ (95\% CI: 0.740--0.746), fall in the range of previously reported structured-only and note-inclusive values on this dataset (Table~\ref{tab:results}).
This is a cross-study comparison rather than a controlled one: the reference cohort is conditioned on discharge-summary availability (303{,}571 vs.\ 422{,}622 admissions) and the absence of per-fold variance precludes a significance test, so we describe our result as competitive with, not superior to, note-based approaches (Section~\ref{sec:limitations}).


\section{Ablation and Comparative Analysis}\label{sec:ablation}

To understand which knowledge sources drive performance, we apply four interpretability methods to LightGBM on B2.
The ablation uses SVD-256 projected B2 features; relative block rankings are consistent with the full-sparse evaluation in Section~\ref{sec:results}.

Table~\ref{tab:ablation} reports the results across eight feature blocks.

\begin{table}[t]
\caption{Feature block importance for LightGBM on B2 (SVD-256, AUROC 0.7399 baseline). Solo: AUROC from one block alone. LOBO (leave-one-block-out): AUROC change when removing one block. Gain: the share of total model gain attributable to a block, where a feature's gain is the summed reduction in training loss across all tree splits that use it, normalized by the total gain over all features. Perm.: AUROC change under random permutation.}\label{tab:ablation}
\centering
\small
\begin{tabular}{lrcccc}
\toprule
\textbf{Feature block} & \textbf{Dims} & \textbf{Solo} & \textbf{LOBO $\Delta$} & \textbf{Gain (\%)} & \textbf{Perm.\ $\Delta$} \\
\midrule
Demographics        & 93     & 0.6985 & $-$0.0369 & 48.5 & $-$0.1117 \\
Organ system labs   & 24     & 0.6273 & $-$0.0019 & 11.4 & $-$0.0227 \\
Drug RxNorm         & 1{,}324  & 0.6576 & $-$0.0041 & 12.8 & $-$0.0192 \\
Unmapped ICD        & 14{,}905 & 0.6569 & $-$0.0040 & 14.4 & $-$0.0142 \\
MONDO L1            & 5{,}413  & 0.6499 & $-$0.0049 &  8.5 & $-$0.0105 \\
CCS procedures      & 527    & 0.6130 & $-$0.0013 &  4.3 & $-$0.0035 \\
MONDO L2            & 892    & 0.5534 & $+$0.0004 &  0.05 & $-$0.0001 \\
MONDO L3            & 252    & 0.5298 & $-$0.0000 &  0.07 & $-$0.0001 \\
\bottomrule
\end{tabular}
\end{table}

\begin{figure}[t]
\centering
\includegraphics[width=0.78\textwidth]{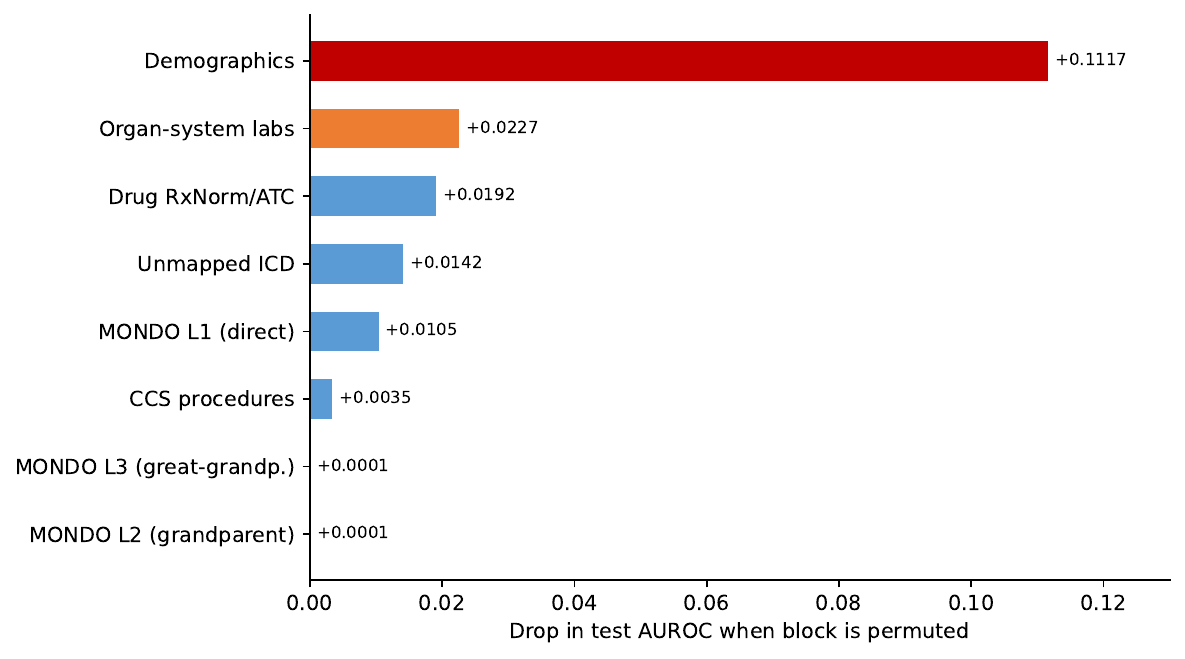}
\caption{Block-level permutation importance for LightGBM on B2. Bar length indicates the drop in test AUROC when the block's features are randomly permuted.}\label{fig:perm_imp}
\end{figure}

\paragraph{Demographics dominate.}
Demographics account for 48.5\% of LightGBM gain and produce the largest permutation importance drop, as shown in Figure~\ref{fig:perm_imp}.
Training on demographics alone yields AUROC 0.6985, comparable to published structured-only baselines such as LACE \cite{van2010derivation} and classical ML on MIMIC-IV \cite{he2025comparative}.

\paragraph{Knowledge source contributions.}
Organ system labs, drug ingredient features, unmapped ICD codes, and MONDO L1 each contribute 1.0 to 2.3 percentage points of permutation importance, with partial redundancy under LOBO removal (Table~\ref{tab:ablation}).
Gain-based importance is highest for unmapped ICD codes and drug ingredient features (Table~\ref{tab:ablation}).
MONDO L2 produces a slight positive LOBO effect and L3 essentially zero; direct disease categories (L1) are therefore sufficient, and broader levels are retained only to preserve the generality of the mapping.

\subsection{Subgroup performance}\label{sec:fairness}
Because demographic blocks dominate model gain (Section~\ref{sec:ablation}), we audit whether discrimination varies across subgroups.
Table~\ref{tab:fairness} reports test-set AUROC with 95\% bootstrap confidence intervals for LightGBM on B2 by sex, insurance, race, and age.

\begin{table}[t]
\caption{Subgroup test-set AUROC for LightGBM on B2 ($n=85{,}040$), with 95\% bootstrap confidence intervals (2{,}000 resamples, fixed seed). Prev.\ (prevalence) is the subgroup 30-day readmission rate. Race categories follow MIMIC-IV groupings, with ``Other/Unk.'' denoting other or unknown race/ethnicity.}\label{tab:fairness}
\centering
\scriptsize
\setlength{\tabcolsep}{4pt}
\renewcommand{\arraystretch}{0.95}
\begin{tabular}[t]{l c c}
\toprule
\textbf{Subgroup} & \textbf{Prev.} & \textbf{AUROC [95\% CI]} \\
\midrule
Overall & 0.202 & 0.741 [0.737, 0.745] \\
\midrule
\textit{Sex} & & \\
\quad Female & 0.189 & 0.746 [0.740, 0.752] \\
\quad Male & 0.215 & 0.734 [0.728, 0.740] \\
\midrule
\textit{Race} & & \\
\quad White & 0.202 & 0.727 [0.722, 0.732] \\
\quad Black & 0.220 & 0.753 [0.743, 0.763] \\
\quad Hispanic & 0.214 & 0.773 [0.758, 0.787] \\
\quad Asian & 0.186 & 0.777 [0.754, 0.798] \\
\quad Other/Unk. & 0.154 & 0.779 [0.763, 0.795] \\
\bottomrule
\end{tabular}
\hfill
\begin{tabular}[t]{l c c}
\toprule
\textbf{Subgroup} & \textbf{Prev.} & \textbf{AUROC [95\% CI]} \\
\midrule
\textit{Insurance} & & \\
\quad Medicare & 0.217 & 0.711 [0.704, 0.718] \\
\quad Medicaid & 0.258 & 0.750 [0.738, 0.762] \\
\quad Other & 0.181 & 0.754 [0.748, 0.760] \\
\midrule
\textit{Age} & & \\
\quad $<$40 & 0.192 & 0.792 [0.783, 0.800] \\
\quad 40--54 & 0.221 & 0.752 [0.744, 0.761] \\
\quad 55--64 & 0.214 & 0.734 [0.724, 0.743] \\
\quad 65--74 & 0.201 & 0.724 [0.714, 0.734] \\
\quad 75--84 & 0.188 & 0.674 [0.662, 0.686] \\
\quad 85+ & 0.164 & 0.668 [0.649, 0.687] \\
\bottomrule
\end{tabular}
\end{table}

Discrimination is stable across sex, insurance, and race, with no subgroup near chance, and is not lower for racial-minority or Medicaid groups.
The main gradient is by age: AUROC declines from 0.79 in the youngest band to 0.67 for patients aged 85 and over, indicating that readmission in the oldest patients depends on factors only partly captured by structured and ontology features.
Comparable AUROC across groups does not imply equitable deployment: at a single operating threshold, false- and true-positive rates vary with subgroup prevalence, and detection is lowest in the oldest patients.
Confidence intervals for the smallest groups are wide and limit interpretation.


\section{Discussion}\label{sec:discussion}

\paragraph{Why ontology features outperform BERT for tree models.}
Gradient-boosted trees partition the feature space through axis-aligned splits, so the sparse ontology features in B2, where each dimension is a single clinical concept, enable semantically meaningful splits, whereas BERT embeddings compress semantics into dense vectors that mix latent factors \cite{grinsztajn2022tree,shwartz2022tabular}.
Conversely, GraphSAGE and LR perform better on B1 (0.3--1.2 AUROC points), benefiting from dense low-dimensional structure for gradient-based optimization; the knowledge-enrichment benefit is thus specific to the model family.

\paragraph{Hierarchy depth and demographic dominance.}
Direct parent-level categories (L1) capture sufficient granularity; broader levels (L2, L3) add no discriminative value (Section~\ref{sec:ablation}). This does not conflict with the benefit of knowledge enrichment: the gain over demographics comes from coarse, interpretable groupings (L1 categories, CCS, organ-system labs, drug ingredients), not deep ontological depth, so a negligible L2/L3 contribution is consistent with a positive overall contribution.
Demographics dominate prediction, reflecting the importance of admission type, insurance status, and length of stay \cite{kum2024systematic}; the four knowledge sources add roughly four AUROC points above demographics alone, though this dominance limits isolation of the knowledge-enrichment contribution.


\section{Limitations and Ethical Considerations}\label{sec:limitations}

\paragraph{Limitations.}
All experiments use single-institution MIMIC-IV, so generalizability to other populations and coding practices is unvalidated.
Cross-validation was run only for LightGBM on B2; other comparisons rely on fixed-split differences without confidence intervals, and the LOBO analysis quantifies block-removal effects but not additive contributions of individual sources.
A configuration combining ontology features with clinical notes was not evaluated; the all-cause label does not separate planned from unplanned readmissions (predominantly emergency and urgent here); and the cross-study comparison with Almeida et al.\ \cite{almeida2025prediction} is qualified by differing inclusion criteria (422{,}622 vs.\ 303{,}571 admissions) and cannot be tested for significance without their per-fold variance.

\paragraph{Ethical considerations and privacy compliance.}
MIMIC-IV is de-identified under the HIPAA Safe Harbor standard and released under a PhysioNet Credentialed Data Use Agreement \cite{benitez2010evaluating,johnson2023mimic}; our pipeline consumes only the de-identified release without re-identification.
Demographic features, including insurance type and race/ethnicity, contribute to model gain and encode socioeconomic determinants that reflect documented systemic health inequities \cite{obermeyer2019dissecting}.
A subgroup audit (Section~\ref{sec:fairness}) shows discrimination is broadly stable across groups, while error rates at a fixed threshold differ by group; deploying such models for resource allocation therefore requires fairness auditing and prospective validation before clinical use.


\section{Conclusion}\label{sec:conclusion}

We proposed a knowledge-enriched feature representation for 30-day readmission prediction that integrates four medical knowledge sources with structured EHR data, without clinical notes.
Under 20-fold cross-validation, LightGBM on B2 achieves AUROC 0.743 without the language-model inference that note-based methods require, using a sparse representation that needs roughly 24 times less storage than the BERT features (Section~\ref{sec:problem}).
Interpretability analysis shows that demographics, together with the knowledge-derived features (direct L1 disease categories, organ-system labs, and drug ingredients), drive prediction, while only the deeper MONDO ontology levels (L2, L3) contribute negligibly.
Overall, enriching structured EHR data with medical knowledge yields an interpretable, lightweight readmission model that performs on par with note-based approaches, without requiring clinical notes.
Future work includes multi-site validation, evaluation of combined ontology and note features, and fairness auditing across demographic subgroups.

\bibliographystyle{splncs04}
\bibliography{references}

\end{document}